# Symbiotic Connectivity: Optimizing Rural Digital Infrastructure with Solar-Powered Mesh Networks Using Multi-Objective Evolutionary Algorithms

Yadira Sanchez Benitez[1,2][0000-0003-0998-3325]

[1] Creative Computing Institute,
University of the Arts London,
London, UK
[2] University of Southampton,
Southampton, UK
`y.sanchezbenitez@arts.ac.uk`

**Abstract.** I present an open-source, ecologically integrated model for rural connectivity, merging the location of nodes mesh networks with renewable energy systems. Employing evolutionary algorithms, this approach optimizes node placement for internet access and symbiotic energy distribution. This model, grounded in community collaboration, demonstrates a balance between technological advancement and environmental stewardship, offering a blueprint for sustainable infrastructure in similar rural settings.

## 1   Introduction

In rural Mexico, like many remote areas worldwide, residents grapple with limited internet and energy supply. Addressing these deficiencies, my research fosters a sustainable infrastructure model that's ecologically considerate and culturally attuned. Following community consultations that identified internet and energy as paramount needs, I will integrate evolutionary algorithms to tailor mesh network node placement and solar infrastructure, emphasizing local culture and ecological harmony.

This approach embodies the essence of evolutionary computation, drawing from the adaptive, innovative mechanisms of biological evolution. Current evolutionary algorithms lack the open-endedness and complexity of biological processes [1], often constrained by small populations and oversimplified mappings. Recognizing these limitations, my work pushes for a nuanced application of these algorithms, capable of mirroring the diversity and resilience of natural systems. In aligning technological expansion with environmental and cultural conservation, the ambition is not just to connect but also to enrich rural communities, catalysing situated infrastructure that is as sustainable as it is inclusive.

## 2    Proposal

I propose an innovative digital infrastructure model that integrates node placement with solar energy deployment in rural communities. This model applies a multi-objective evolutionary algorithm (MOEA) to optimize network and energy distribution, factoring in the local topography and housing density to ensure ecological and cultural sensitivity. Multi-objective evolutionary algorithms have been used in water distribution network design[2], in the design and sizing stage of pedestrian bridges[3] and energy management of agricultural systems[4]. However, there is no application of multi-objective evolutionary algorithms in the context proposed here. Therefore, the algorithm I propose here allows for a nuanced approach, sidestepping the limitations of small population models and direct mappings, instead offering a resilient and adaptable network reflective of the rural community's unique landscape and stated needs.

The pseudo algorithm in Table 1 describes the overall structure.

**Table 1.** Algorithm: Solar-Powered Node Placement Tool

```
                      PARAMETERS:
                   Current iteration, itr
                   Iteration limit, imax
              Solar data refresh interval, sdr
                Energy consumption model, ECM
                Energy production model, EPM

                         INPUT:
                    Node locations, N
                   Solar energy map, SEM
                        OUTPUT:
                Optimal node placement, NP
```

```
1: G ← geographic area for node placement
2: N ← initial set of node locations
3: NP ← N with minimum energy deficit
4: Ps ← secondary node placement options
5: Pf ← final set of node placement options
6: Generate G based on SEM
7: while itr ≤ imax do
8: Initialise Pn based on ECM and EPM
9: Evaluate and sort N based on energy balance
10: function SolarEvaluation(N)
11: SE ← Energy balance after solar evaluation
12: end function
13: function EnergyOptimization(N)
14: EO ← Node placement after energy optimization
15: end function
16: Generate Ps using SE and EO
17: Pf ← NP ∪ SE ∪ EO ∪ Ps
18: Evaluate Pf and find NP with the best energy balance
19: if itr mod sdr == 0 then
20: Refresh SEM
```

```
21: Recalculate energy balance for NP
22: end if
23: itr ← (itr + 1)
24: end while
return NP
```

The pseudocode I have outlined responds to the requirements of my approach. It establishes a procedural foundation for integrating node placement with solar energy deployment.

The algorithm uses G as the geographic area for node placement, with N representing the initial set of nodes. NP denotes optimal node placements considering solar power, while Ps and Pf are secondary and final node placement options. It generates G based on a Solar Energy Map (SEM), initializing Pn with an Energy Consumption Model (ECM) and Energy Production Model (EPM). SolarEvaluation and EnergyOptimization functions evaluate and optimize node placement based on solar energy availability. The algorithm refreshes SEM periodically and iterates until reaching the limit to find the best energy balance in NP.

## 3   Work Progress

The ongoing work represents a crucial step towards harnessing the full capabilities of the proposed algorithm. By aligning with steps 1 to 5 of the pseudocode, this initiative not only integrates topographical and housing density data for node placement simulation but also directly addresses the pressing needs of rural communities. Emphasizing equitable and grounded solutions, the project aims to bridge the digital divide by tailoring infrastructure to meet the actual requirements of the population, thus fostering more inclusive and responsive development[5].

Figure 1 shows a simulation node placement generated by the tool.

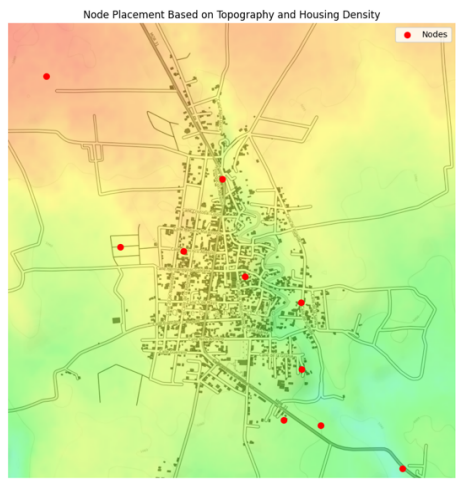

**Fig. 1: An example of node placement based on topography and house density**

The red dots represent the nodes positioned in accordance with the topography and housing density, offering an initial visualization of the potential network layout optimized for solar energy usage. This work sets the stage for a more sophisticated analysis that will incorporate a multi-objective evolutionary algorithm (MOEA) to fine-tune node distribution in harmony with solar infrastructure placement, aiming for a sustainable and efficient digital infrastructure design.

## 4     Conclusion and Future Work

Future endeavors will focus on implementing the more intricate components (steps 6 to 24) of the algorithm, incorporating solar evaluation and energy optimization functions. This advancement aims to align node placements not only with network performance but also with efficient solar energy utilization, fostering a symbiotic relationship between connectivity and sustainable energy in the community's infrastructure.

## References

1. Miikkulainen, R., Forrest, S.: A biological perspective on evolutionary computation. Nature Machine Intelligence. 3, 9–15 (2021).
2. Kidanu, R.A., Cunha, M., Salomons, E., Ostfeld, A.: Improving multi-objective optimization methods of Water Distribution Networks. Water. 15, 2561 (2023).
3. Tres Junior, F.L., Yepes, V., Medeiros, G.F., Kripka, M.: Multi-objective optimization applied to the design of Sustainable Pedestrian Bridges. International Journal of Environmental Research and Public Health. 20, 3190 (2023).
4. Shamshirband, S., Khoshnevisan, B., Yousefi, M., Bolandnazar, E., Anuar, N.B., Abdul Wahab, A.W., Khan, S.U.: A multi-objective evolutionary algorithm for Energy Management of Agricultural Systems—a case study in Iran. Renewable and Sustainable Energy Reviews. 44, 457–465 (2015).
5. Sanchez Benitez, Y.: Algorithmic ecologies of justice: Using computational social science methods to co-design tools of resistance, resilience and care with communities. 13th ACM Web Science Conference 2021. (2021).